\pdfoutput=1

\documentclass[11pt]{article}

\usepackage{acl}

\usepackage[T1]{fontenc}
\usepackage{floatrow}
\usepackage{kotex}
\usepackage{makecell}
\usepackage{caption}
\usepackage{array}
\usepackage{capt-of}
\usepackage{ulem}
\usepackage{longtable}
\usepackage[utf8]{inputenc}
\usepackage{authblk}
\usepackage{enumitem}
\usepackage{amsmath}
\DeclareMathOperator*{\argmax}{arg\,max}
\usepackage{float}
\usepackage{times}
\usepackage{latexsym}
\usepackage{placeins}
\usepackage{graphicx}
\usepackage{stackengine}

\graphicspath{ {/images/} }

\usepackage[T1]{fontenc}

\usepackage[utf8]{inputenc}

\usepackage{microtype}
\usepackage{inconsolata}
\setlist[itemize]{noitemsep, topsep=2pt}

\usepackage{graphicx}               
\usepackage{tabularx}               
\usepackage{booktabs}
\usepackage{amsmath}
\usepackage{stmaryrd}
\usepackage{xcolor}
\usepackage{colortbl}
\usepackage{soul}

\newcolumntype{C}{>{\centering\arraybackslash}X}
\usepackage{multirow}               
\usepackage{diagbox}                
\usepackage{hhline}                 
\usepackage{color}                  
\usepackage{titlesec}
\usepackage{amsmath}                
\usepackage{amssymb}                
\usepackage{mathtools}              
\usepackage{enumitem}               

\usepackage{subfigure}
\usepackage{booktabs}
\usepackage{extarrows}
\usepackage{makecell}
\usepackage{tablefootnote}
\usepackage{soul}

\usepackage{amsmath,amsfonts,bm}

\def\eqref#1{equation~\ref{#1}}

\def\1{\bm{1}}

\DeclareMathAlphabet{\mathsfit}{\encodingdefault}{\sfdefault}{m}{sl}
\SetMathAlphabet{\mathsfit}{bold}{\encodingdefault}{\sfdefault}{bx}{n}

\newcommand{\sref}[1]{\S\ref{#1}} %

\newcommand{\cmc}{\texttt{CMC}}

\newcommand{\bi}{bi-encoders}
\newcommand{\ce}{cross-encoders}
\newcommand{\biint}{BE}
\newcommand{\ceint}{CE}

\definecolor{RoseQuartzBg}{HTML}{F7CAC9}
\definecolor{RoseQuartz}{HTML}{F5A798}
\definecolor{Serenity}{HTML}{92A8D1}
\definecolor{OrangeRed}{rgb}{1.0, 0.27, 0.0}
\definecolor{Red}{rgb}{1.0, 0.0, 0.0}
\definecolor{Turquoise}{HTML}{0F4C81}
\usepackage{xparse}
\NewDocumentCommand{\nishant}{ mO{} }{\textcolor{blue}{\textsuperscript{\textit{Nishant}}\textsf{\textbf{\small[#1]}}}}
\NewDocumentCommand{\wenlong}{ mO{} }{\textcolor{Serenity}{\textsuperscript{\textit{Wenlong}}\textsf{\textbf{\small[#1]}}}}
\NewDocumentCommand{\jy}{ mO{} }{\textcolor{RoseQuartz}{\textsuperscript{\textit{jy}}\textsf{\textbf{\small[#1]}}}}
\NewDocumentCommand{\jyc}{ mO{} }{\textcolor{blue}{\textsuperscript{\textit{jy}}\textsf{\textbf{\small[#1]}}}}
\NewDocumentCommand{\js}{ mO{} }{\textcolor{OrangeRed}{\textsuperscript{\textit{jong}}\textsf{\textbf{\small[#1]}}}}
\NewDocumentCommand{\jsc}{ mO{} }{\textcolor{blue}{\textsuperscript{\textit{jong comment}}\textsf{\textbf{\small[#1]}}}}

\title{Comparing Neighbors Together Makes it Easy: Jointly Comparing Multiple Candidates for Efficient and Effective Retrieval}

\author{\textbf{Jonghyun Song}$^\dag$, \textbf{Cheyon Jin}$^\dag$, \textbf{Wenlong Zhao}$^{\diamondsuit}$, \textbf{Andrew McCallum}$^{\diamondsuit}$, \textbf{Jay-Yoon Lee\thanks{Corresponding author}}$^\dag$ \\
$^\dag$ Seoul National University \quad $^{\diamondsuit}$University of Massachusetts Amherst \\
\texttt{\{hyeongoon11, cheyonjin, lee.jayyoon\}@snu.ac.kr} \\ \quad \texttt{\{wenlongzhao, mccallum\}@umass.edu}
}

\begin{document}
\maketitle
\begin{abstract}
A common retrieve-and-rerank paradigm involves retrieving relevant candidates from a broad set using a fast bi-encoder (BE), followed by applying expensive but accurate cross-encoders (CE) to a limited candidate set.  However, relying on this small subset is often susceptible to error propagation from the bi-encoders, which limits the overall performance. To address these issues, we propose the Comparing Multiple Candidates (\cmc) framework. \cmc\ compares a query and multiple embeddings of similar candidates (i.e., neighbors) through shallow self-attention layers, delivering rich representations contextualized to each other. Furthermore, \cmc\ is scalable enough to handle multiple comparisons simultaneously. For example, comparing ~10K candidates with \cmc\ takes a similar amount of time as comparing 16 candidates with CE. Experimental results on the ZeSHEL dataset demonstrate that \cmc, when plugged in between bi-encoders and cross-encoders as a seamless intermediate reranker (BE-CMC-CE), can effectively improve recall@k (+4.8\%-p, +3.5\%-p for R@16, R@64) compared to using only bi-encoders (BE-CE), with negligible slowdown (<7\%). Additionally, to verify \cmc's effectiveness as the final-stage reranker in improving top-1 accuracy, we conduct experiments on downstream tasks such as entity, passage, and dialogue ranking. The results indicate that \cmc\ is not only faster (11x) but also often more effective than \ce\, with improved prediction accuracy in Wikipedia entity linking (+0.7\%-p) and DSTC7 dialogue ranking (+3.3\%-p). 
\end{abstract}

\section{Introduction}
\begin{figure*}[hbt!]
\centering    \includegraphics[width=0.92\textwidth]{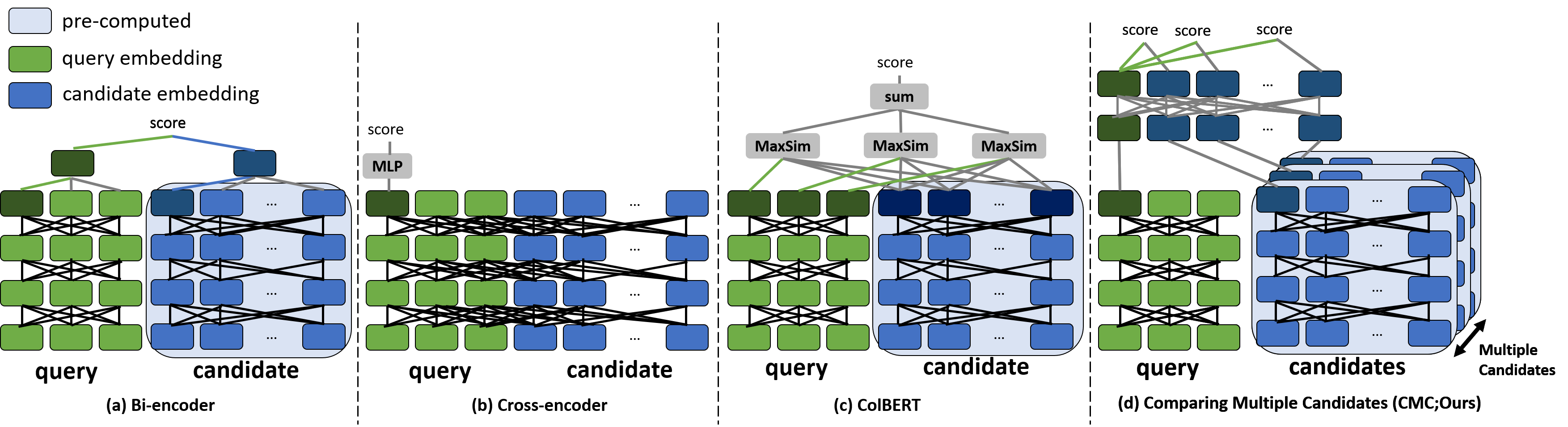}
\caption{Model architectures for retrieval tasks. (a), (b), and (c) are existing architectures. (d) is our proposed `Comparing Multiple Candidates (\cmc)' architecture, which computes compatibility score by comparing the embeddings of a query and K multiple candidates via self-attention layers. Contrary to (a)-(c), \cmc\ can process multiple candidates at once rather than conducting several forward passes for each (query, candidate) pair.}
    \label{fig:score}
\end{figure*}

The two-stage approach of retrieval and reranking has become a predominant method in tasks such as entity linking (EL) \cite{wu2020scalable, zhang2021understanding, xu2023read},  open-domain question answering (ODQA) \cite{nogueira2019passage, agarwal2022zero, shen2022low, qu2020open}, and dialogue systems \cite{mele2020topic}.
Typically, \bi\ (\biint) are used to efficiently retrieve relevant candidates from a large set of documents (e.g., knowledge base), and then \ce\ (\ceint)  effectively rerank only a limited subset of candidates already retrieved by \biint\ (\citet{nogueira2019passage}; Figure \ref{fig:score}.a-b).

The current BE-CE approach, although widely adopted, has an efficiency-effectiveness trade-off and is susceptible to error propagation. When less accurate \biint\ retrieves candidates, the whole framework risks the error propagation of missing the gold candidates due to inaccuracies from the retriever. Simply increasing the number of candidates is not a viable solution considering the slow serving time of \ceint\footnote{For the serving time of cross-encoders, see \sref{sec:app-runtime}.}\footnote{Furthermore, increasing the number of candidates for \ceint\ does not necessarily improve end-to-end accuracy \cite{wu2020scalable}. We confirm this in the experiments. See appendix \ref{subsec:cross-num}.}.
Consequently, users are faced with the dilemma of deciding which is worse: error propagation from \biint\ versus the slow runtime of \ceint.

To resolve this issue, various strategies have been proposed to find an optimal balance in the efficiency-effectiveness trade-off. Prior works (\citet{khattab2020colbert, zhang2021understanding, cao-etal-2020-deformer, humeau2019poly}) have enhanced bi-encoder architectures with a late interaction component. 
However, these models only focus on single query-candidate pair interaction. Also, they sometimes require storing entire token embeddings per candidate sentence which results in tremendous memory use (Figure \ref{fig:score}.c). 

Our proposed Comparing Multiple Candidates (\cmc) makes reranking easy by comparing similar candidates (i.e., neighbors) together. By jointly contextualizing the single vector embeddings from each candidate through shallow bi-directional self-attention layers, \cmc\ achieves high prediction accuracy and runtime efficiency that are comparable to, or better, than existing methods which require single or multiple vector embeddings.

In other words, \cmc\ only takes a single forward pass for input $(\text{query}, \text{candidate}_1, ... , \text{candidate}_k)$ with a pre-computed single vector embedding. In contrast, models such as \ceint\ and other late interaction models take $k$ separate forward passes for input pairs $(\text{query}, \text{candidate}_1), ... ,(\text{query}, \text{candidate}_k)$, sometimes requiring multiple vector embeddings per each candidate. \cmc\ maintains both the \textit{efficiency} of \biint\ with pre-computed single-vector candidate embeddings, and the \textit{effectiveness }of \ceint\ with interactions between query and multiple candidates (Figure \ref{fig:score}.d).



\begin{figure*}[hbt!]
\centering
    \includegraphics[width=0.97\textwidth]{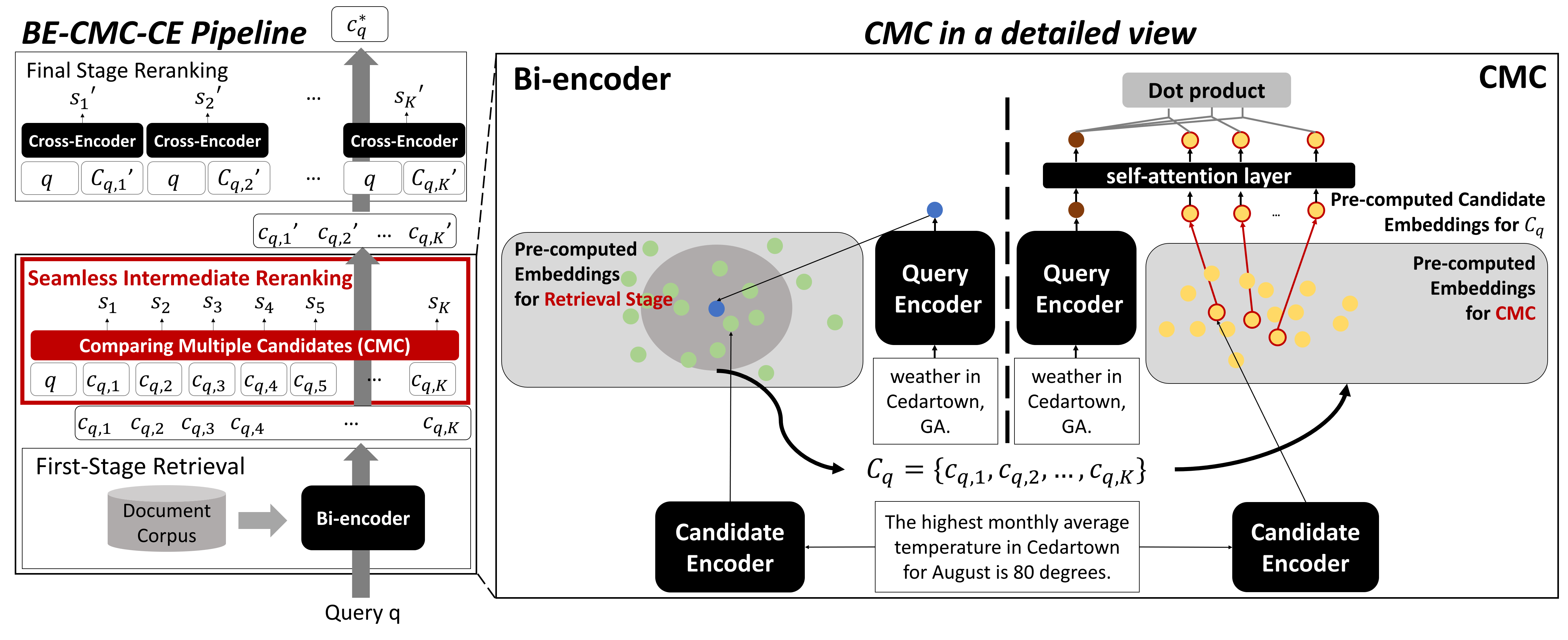}
\caption{Overview of the proposed \textsc{\cmc} framework that compares multiple candidates at once. \cmc\ can \textit{seamlessly enhance retriever}, finding top-K' candidates, or function as a direct reranker which outputs top-1 candidate. Candidate embeddings for \bi\ and \cmc\ are both precomputed while query embeddings for \bi\ and \cmc\ are computed in parallel on the fly. After \bi\ retrieve top-$K$ candidates, \cmc\ indexes the corresponding candidate embeddings and passes through a two-layer transformer encoder. Here, the additional latency is limited to the execution of self-attention layers.}
    \label{fig:description}
\end{figure*}

Practitioners can plug in \cmc\ as the \textit{seamless intermediate reranker} (BE-$\cmc$-CE) which can enhance retrieval performance with negligible extra latency.
This improvement is crucial for preventing error propagation from the retrieval process, resulting in more reliable candidates for the final stage (Figure \ref{fig:description}-\ref{fig:be-cmc-ce}). On the other hand, \cmc\ also can serve as a fast and effective \textit{final-stage reranker} improving top-1 accuracy (BE-$\cmc$). If there’s a time constraint, using \cmc\ as the final reranker can be a good option, as running a cross-encoder requires significantly more time (Table \ref{tab:ranking}; Figure \ref{fig:runtime}).


In experiments, we evaluate \cmc\ on Zero-SHot Entity-Linking dataset (ZeSHEL; \citet{logeswaran2019zero}) to investigate how much \cmc\ seamlessly enhances a retriever's performance when plugged in to BE (BE-CMC). The results show \cmc\ provides higher recall than baseline retrievers at a marginal increase in latency (+0.07x; Table \ref{tab:retriever}). Compared to standard \biint-\ceint, plugging in \cmc\ as the seamless intermediate reranker (\biint-CMC-\ceint) can provide fewer, higher-quality candidates to \ceint, ultimately improving the accuracy of  \ceint\ reranking. (Table \ref{tab:retriever-ce}). To examine the effectiveness of \cmc\ which acts as the final stage reranker, we evaluate \cmc\ on entity, passage, and dialogue ranking tasks. We observe that \cmc\ outperforms \ceint\ on Wikipedia entity linking datasets (+0.7p accuracy) and DSTC7 dialogue ranking datasets (+3.3p MRR), requiring only a small amount (0.09x) of \ceint 's latency (Table \ref{tab:ranking}).

The main contributions of the paper are as follows:
\begin{itemize}[noitemsep, topsep=2pt]
\item  We present a novel reranker, \cmc, which improves both accuracy and scalability. \cmc\ contextualizes candidate representations with similar candidates (i.e., neighbors), instead of solely focusing on a single query-candidate pair (\sref{sec:method}).

\item \cmc\ can serve as the seamless intermediate reranker which can significantly improve retrieval performance with only a negligible increase in latency. This results in a more confident set of candidates for the final-stage reranker that improves end-to-end accuracy compared to conventional bi-encoders (\sref{subsec:retriever})
\item Experimental results show that the final stage reranking of \cmc\ is highly effective on passage, entity, and dialogue ranking tasks compared to various baselines among the low-latency models (\sref{subsec:reranker}).
\item Additionally, we show that \cmc\ can benefit from domain transfer from sentence encoders while \biint\ and many others cannot (\sref{subsec:ablation}).
\end{itemize}

\section{Background and Related Works}



\label{subsec:related}
\subsection{Retrieve and Rerank}
Two-stage retrieval systems commonly consist of a fast retriever and a slow but accurate reranker. 
Although the retriever is fast, its top-1 accuracy tends to be suboptimal. Therefore, a candidate set $C_q= \{c_{q,1}, c_{q,2}, \ldots, c_{q,K}\} \subseteq \mathcal{C}$  whose elements are $K$ most relevant candidates in the corpus $\mathcal{C}$ is retrieved for further reranking.

A reranker $s_\theta(q, c_{q,j}) (1\leq j\leq K)$ is a model trained to assign a fine-grained score between the query $q$ and each candidate $c_{q,j}$ from the relatively small set of candidates $C_q$. 
It is an expressive model that is slower but more accurate than the retriever. 
The candidate with the highest score $c_q^* = \argmax_{c_{q,j} \in C_q} s_\theta(q, c_{q,j})$
is the final output of the retrieve-and-rerank pipeline where query $q$ should be linked.
\subsection{Related Work}
\paragraph{Bi-encoders and Cross-encoders}
In two-stage retrieval, the compatibility score between the query and candidate can be computed by diverse functions. \citet{nogueira2019multi} retrieve candidates using the bag-of-words BM25 retriever and then apply a cross-encoder reranker, transformer encoders that take the concatenated query and candidate tokens as input \cite{logeswaran2019zero, wu2020scalable}. Instead of BM25 retriever, other works \cite{lee2019latent, gillick2019learning,karpukhin2020dense} employ a pre-trained language model for a bi-encoders retriever to encode a query and a candidate separately and get the compatibility score.
The scalability of bi-encoders as a retriever arises from the indexing of candidates and maximum inner-product search (MIPS); however, they tend to be less effective than cross-encoders as candidate representations do not reflect the query's information (Figure \ref{fig:score}.a-b). To enhance the performance of bi-encoders, follow-up works propose a task-specific fine-tuned model \cite{gao2022unsupervised}, injecting graph information \cite{wu2023modeling, agarwal2022entity}, and multi-view text representations \cite{ma2021muver, liu2023towards}. 
\paragraph{Late Interaction}
Late interaction models, which typically function as either a retriever or a reranker, enhance bi-encoder architectures with a late interaction component between the query and the candidate.

Poly-encoder \cite{humeau2019poly} and MixEncoder \cite{yang-etal-2023-enough} represent query information through cross-attention with a candidate to compute the matching score. However, these models have overlooked the opportunity to explore the interaction among candidates. 

Sum-of-Max \cite{khattab2020colbert, zhang2021understanding} and DeFormer \cite{cao-etal-2020-deformer} rely on maximum similarity operations or extra cross-encoder layers on top of bi-encoders. However, they lack scalability due to the need to pre-compute and save every token embedding per each candidate.\footnote{For example, 3.2TB is required for storing $\sim$5M entity descriptions from Wikipedia, each with 128 tokens. In contrast, storing a single vector embedding per entity description for bi-encoders only requires 23GB.} As a collection of documents continuously changes and grows, this storage requirement poses practical limitations on managing and updating the document indices.

\cmc\ differs from these models in its enhanced scalability by comparing a single embedding for each candidate. This approach provides a deeper exploration of relational dynamics from interactions across multiple candidates while improving time and memory efficiency. 

\paragraph{Listwise Ranking}
\cmc\ is not the first approach to compare a list of documents to enhance ranking performance  \cite{han2020learning, zhang2022hlatr, xu2023read}. These listwise ranking methods process cross-encoder logits for the list $(\text{query}, \text{candidate}_1, \ldots, \text{candidate}_\text{K})$ to rerank $K$ candidates from cross-encoders. However, these approaches lack scalability and efficiency due to reliance on cross-encoder representations.

Unlike previous listwise ranking models, we propose a method that employs representations from independent sentence encoders rather than cross-encoders. Boosting scalability with independent representations, \cmc\ can seamlessly enhance retrievers by maintaining prediction accuracy.

\section{Proposed Method}
\label{sec:method}
\subsection{Model Architecture}
Comparing Multiple Candidates, \cmc, employs shallow self-attention layers to capture both query-candidate and candidate-candidate interactions.
Unlike other late interaction models which compute the compatibility scores by only considering a single query-candidate pair \cite{khattab2020colbert, humeau2019poly, yang-etal-2023-enough}, \cmc\ compares each candidate to the query and other candidates at the same time (Figure \ref{fig:score}.(d)).
The self-attention layer in \cmc\ processes the concatenated representations of the query and multiple candidates, derived from the independent query and candidate encoders.
In this way, \cmc\ obtains enhanced representations of the query and every candidate by contextualizing them with each other. Also, this architecture is scalable to a large set of corpus by pre-computing and indexing candidate embeddings. For example, processing 2K candidates only takes twice as long as processing 100 (Figure \ref{fig:runtime}).

\paragraph{Query and Candidate Encoders}
Prior to \cmc, the first-stage retriever (e.g., bi-encoders) retrieves the candidate set with K elements  \( C_q=\{c_{q,1},...c_{q,K}\} \) for query $q$.  \cmc\ then obtains the aggregated encoder output (e.g., \texttt{[CLS]} token embedding) of query sentence tokens  $\mathbf{h}^{sent}_q$ and candidate sentence tokens $\mathbf{h}^{sent}_{c_{q,j}}$ from the query encoder \( \texttt{Enc}_{qry} \) and the candidate encoder \( \texttt{Enc}_{can} \). These encoders play the same role as conventional bi-encoders by condensing each query and candidate information into a single vector embedding but are trained separately from the first-stage stage retriever. 
\begin{align}
\mathbf{h}^{sent}_{q} &= \texttt{agg}( \texttt{Enc}_{qry} ([\texttt{CLS}]\text{x}^{0}_{q}\ldots \text{x}^{k}_{q})) \\
\mathbf{h}^{sent}_{c_{q,j}} &= \texttt{agg}( \texttt{Enc}_{can} ([\texttt{CLS}]\text{x}^{0}_{c_{q,j}}\ldots \text{x}^{k}_{c_{q,j}}))&&
\end{align}
\( \text{x}_q \) and \( \text{x}_{c_{q,j}} \) are tokens of each query and candidate. The aggregator function \texttt{agg} extracts \texttt{[CLS]} embedding from the last layer of encoder\footnote {For entity linking tasks, both the query (mention) and candidate (entity) sentences include custom special tokens that denote the locations of mention and entity words. These include $\texttt{[SEP]}$, $\texttt{[query\_start]}$, $\texttt{[query\_end]}$, and $\texttt{[DOC]}$ tokens following \citet{wu2020scalable}.}.
\paragraph{Self-attention Layer}
The shallow self-attention layers process concatenated embeddings of a query and all candidates. This lightweight module enables parallel computation (\textit{efficient}) and outputs contextualized embeddings via interactions between query and candidates (\textit{effective}). In the reranking perspective, Representing candidates together with self-attention layers (\texttt{Attn}) enables fine-grained comparison among candidates. The self-attention layers consist of two layers of vanilla transformer encoder \cite{vaswani2017attention} in Pytorch without positional encoding. 
\begin{equation}
\resizebox{\columnwidth}{!}{%
$\left[\mathbf{h}^{\cmc}_{q}; \mathbf{h}^{\cmc}_{c_{q, 1}}; \ldots; \mathbf{h}^{\cmc}_{c_{q, K}}\right] = \texttt{Attn}\left( \left[\mathbf{h}^{sent}_{q}; \mathbf{h}^{sent}_{c_{q, 1}}; \ldots; \mathbf{h}^{sent}_{c_{q, K}}\right] \right)$%
}
\end{equation}

Subsequently, the reranker computes the final prediction \( c^*_{q} \) via dot products of query and candidate embeddings from the self-attention layer:
\begin{equation}
c^*_{q} = \argmax_{c_{q, j} \in C_{q}}  \mathbf{h}^{\cmc}_{q} \cdot\left( \mathbf{h}^{\cmc}_{c_{q, j}} \right)^\top 
\end{equation}
\subsection{Training}
\label{sec:training}
\paragraph{Optimization}
The training objective is minimizing the cross-entropy loss regularized by the Kullback-Leibler (KL) divergence between the score distribution of the trained model and the bi-encoder. The loss function is formulated as:
\begin{equation}
\resizebox{\columnwidth}{!}{$
\mathcal{L}(q, \tilde{C}_q) = \sum_{i=1}^{K} (-\lambda_1 y_i \log(p_i) +\lambda_2  p_i \log\left(\frac{p_i}{r_i}\right))
\label{eq:loss}
$}
\end{equation}

\( y_i \) and \( p_i \) are the ground truth and predicted probability for i-th candidate. The retriever's probability for the candidate is represented as \( r_i \). \(\lambda_1\) and \(\lambda_2\) are weights combining the two losses.

\paragraph{Negative Sampling}
We sample hard negatives based on the first-stage retriever's score for each query-candidate pair $(q, c_{q,j})$: $\forall j \in \{1, \ldots, K\} \setminus \{\text{gold index}\}$, 
\begin{equation}
\quad {c}_{q, j} \sim \frac{\exp(s_{\text{retriever}}(q, c_{q, j}))}{\sum_{\substack{k=1 \&\\ k \neq \text{gold index}}}^K \exp(s_{\text{retriever}}(q, c_{q, k}))}
\label{eq:sampling}
\end{equation}
\newline
In experiments, \cmc\ and other baselines follow the same optimization and negative sampling strategy.\footnote{The code and link to datasets are available at https://github.com/yc-song/cmc}

\subsection{Inference}
\label{subsec:inference}
\paragraph{Offline Indexing} \cmc\ can pre-compute and index the embeddings of candidates in the collection (e.g., knowledge base), unlike cross-encoders (Figure~\ref{fig:score}). This offline indexing scheme significantly reduces inference time compared to cross-encoders, making the runtime of \cmc\ comparable to that of bi-encoders (\sref{subsec:reranker}). While reducing time complexity, \cmc\ is highly memory-efficient requiring less than $1\%$ of index size needed by Sum-of-Max and Deformer, which store every token embedding per candidate. This is because \cmc\ only stores a single vector embedding per candidate.

\paragraph{Parallel Computation of Query Representations}
\label{subsec:parallel}
The end-to-end runtime for retrieving and reranking with \cmc\ can be comparable to that of bi-encoder retrieval. The runtime can be further improved by parallelizing query encoders in both bi-encoder and \cmc\ (Figure~\ref{fig:description}). Ideally, the additional latency for running \cmc\ is limited to the execution of a few self-attention layers.

\paragraph{\cmc\ as the Seamless Intermediate Reranker} 
\cmc\ can serve as a seamless intermediate reranker that
maintains the latency-wise user experience while providing improved retrieval performance when combined with a bi-encoder. Thanks to the parallel computation discussed above, plugging in CMC after bi-encoders should minimally impact retrieval latency compared to just using the bi-encoder. The process starts with the first-stage retrievers, such as bi-encoders, retrieving a broad set of candidates. \cmc\ then narrows this set down to fewer, higher-quality candidates with a more manageable number (e.g., 64 or fewer) for the reranker. Since \cmc , the seamless intermediate reranker, filters candidates from the first-stage retriever with negligible additional latency, its runtime is comparable to that of bi-encoders. As a result, the improved candidate quality boosts the prediction accuracy of the final-stage reranker (e.g., cross-encoders) with only a marginal increase in computational cost (Figure \ref{fig:be-cmc-ce}; \sref{subsec:retriever}).

\paragraph{\cmc\ as the Final Stage Reranker} \cmc\ can obviously serve as the final-stage reranker to increase top-1 accuracy. Enriching contextualized representations of the query and candidates helps improve top-1 accuracy in reranking while maintaining efficiency with a single vector embedding. Notably, \cmc\ remains effective even when the number of candidates varies during inference, despite being trained with a fixed number of candidates. For example, when trained with 64 candidates on the MS MARCO passage ranking dataset, \cmc\ still performs effectively with up to 1K candidates. This demonstrates not only the scalability of \cmc\ but also its robustness in processing a diverse range of candidate sets (\sref{subsec:reranker}).

\section{Experiments}
\begin{table*}[!ht]
\centering
\resizebox{0.9\textwidth}{!}{
\begin{tabular}{ll|llllll|ll}
\hline
        &                                        & \multicolumn{6}{c|}{Test}                                                                                             & Speed& Index Size\\
        & Method                                 & R@1               & R@4               & R@8               & R@16              & R@32              & R@64              & (ms)& (GB)\\ \hline
Single- & BM25         & 25.9               & 44.9                 & 52.1                 & 58.2                 & 63.8                 & 69.1             & & \\
View         & Bi-encoder (BE$^\spadesuit$)& 52.9             & 64.5             & 71.9             & 81.5             & 85.0             & 88.0             & 568.9& 0.2\\
        & Arbo-EL        & 50.3             & 68.3             & 74.3             & 78.4              & 82.0             & 85.1             & -                 & -                  \\
        & GER                         & 42.9             & 66.5             & 73.0             & 78.1             & 81.1             & 85.7             & -              & -              \\
        & Poly-encoder (Poly) $^\heartsuit$                        & 40.0\small$\pm 0.7$           & 60.2\small$\pm 0.9$             & 67.2\small$\pm 0.7$             & 72.2\small$\pm 0.8$             & 76.5\small$\pm 0.8$             & 80.2\small$\pm 0.8$             & 581.0& 0.2\\
        & BE + Poly$^\heartsuit$ & 56.9\small$\pm 0.8$& 74.8\small$\pm 0.6$& 80.1\small$\pm 0.7$& 84.2\small$\pm 0.5$& 87.5\small$\pm 0.4$& 90.2\small$\pm 0.3$& 574.6& 0.4\\ 
        & Sum-of-max (SOM)$^\heartsuit$                              & 27.1\small$\pm 1.8$             & 64.1\small$\pm 1.4$             & 73.2\small$\pm 0.9$             & 79.6\small$\pm 0.7$             & 84.1\small$\pm 0.4$             & 88.0\small$\pm 0.4$             & 6393.0& 25.7\\
        & BE + SOM$^\heartsuit$ & \multirow{2}{*}{\underline{58.5}\small$\pm 1.0$}& \multirow{2}{*}{\underline{76.2}\small$\pm 1.1$}& \multirow{2}{*}{\underline{81.6}\small$\pm 1.0$}& \multirow{2}{*}{\underline{85.8}\small$\pm 0.9$}& \multirow{2}{*}{\underline{88.9}\small$\pm 0.7$}& \multirow{2}{*}{\underline{91.4}\small$\pm 0.6$}& 2958.3& 0.2\\ 
        & - w/ offline indexing & & & & & & & 597.3& 25.9\\ 
        & BE$^\spadesuit$ + \cmc (Ours)& \textbf{59.1\small$\pm 0.3$}             & \textbf{77.6\small$\pm 0.3$}             & \textbf{82.9\small$\pm 0.1$}    & \textbf{86.3\small$\pm 0.2$}    & \textbf{89.3\small$\pm 0.2$}    & \textbf{91.5\small$\pm 0.1$}    & 607.2& 0.4 \\
        \hline
Multi-  & MuVER                & 43.5             & 68.8             & 75.9             & 77.7             & 85.9             & 89.5             & -                 & -                  \\
View    & MVD               & \underline{52.5} & \underline{73.4} & \underline{79.7} & \underline{84.4} & \underline{88.2} & \underline{91.6} & -                 & -                  \\
        & MVD + \cmc (Ours)                    & \textbf{59.0}    & \textbf{77.8}    & \textbf{83.1}    & \textbf{86.7}    & \textbf{89.9}    & \textbf{92.4}    & -& -\\ \hline
\end{tabular}%
}
\caption{Retrieval performance over ZeSHEL dataset. The best and second-best results are denoted in bold and underlined. \biint$^\spadesuit$ is bi-encoder from \citet{yadav2022efficient} which is used for \cmc. $^\heartsuit$ indicates our implementation as recall@k for all k are not provided in previous work\protect\footnotemark. results on BE + Reranker (e.g., BE+\cmc) are conducted over the top 512 candidates from the first-stage retriever and averaged over experiments with 5 random seeds.}
\label{tab:retriever}
\end{table*}
\footnotetext{recall@64 of Poly-encoder and Sum-of-max from \citet{zhang2021understanding} is reported as 84.34 and 89.62, respectively.}

\subsection{Dataset}
To evaluate the robustness of \cmc, we conduct experiments on various ranking tasks where the retrieve-and-rerank approach is commonly used. For entity linking, we utilize datasets linked to the Wikipedia knowledge base (AIDA-CoNLL \cite{hoffart2011robust}, WNED-CWEB \cite{guo2018robust}, and MSNBC \cite{cucerzan2007large}), as well as a ZEro-SHot Entity Linking dataset (ZeSHEL; \citet{logeswaran2019zero}) based on the Wikia\footnote{now Fandom: \url{https://www.fandom.com}} knowledge base. The candidates are retrieved from bi-encoders fine-tuned for each knowledge base \cite{wu2020scalable, yadav2022efficient}. For passage ranking, we conduct an experiment on MS MARCO with 1K candidates from BM25 as the first-stage retriever following \citet{nguyen2016ms}. For dialogue ranking tasks, we test our model on DSTC7 challenge (Track 1) \cite{yoshino2019dialog}, where candidates are officially provided. The primary metric used is recall@k, as datasets typically have only one answer or rarely a few answers per query. Further details are presented in \sref{sec:stat-datasets}.

\subsection{Training Details}
\cmc\ and other baselines are trained under the same training strategies. All models use the same loss function and negative sampling (\sref{sec:training}) with the AdamW optimizer and a 10\% linear warmup scheduler. Also, we examine diverse sentence encoder initialization for \cmc\ and late interaction models, including vanilla BERT and BERT-based models fine-tuned on in- and out-of-domain datasets. After training, we select the best results for each model.\footnote{If more favorable results are found in prior works over the same candidates, we use those results.} For ZeSHEL, training \cmc\ and other low-latency baselines for one epoch on an NVIDIA A100 GPU takes about 4 hours. The training details for each dataset are in \sref{sec:detail-training}, and the ablation studies for diverse training strategies are presented in \sref{subsec:ablation} and \sref{subsec:ablation-training}.

\subsection{CMC as the Seamless Intermediate Reranker}
\label{subsec:retriever}

\begin{figure}[t!]
  \centering
  \includegraphics[width=\columnwidth]{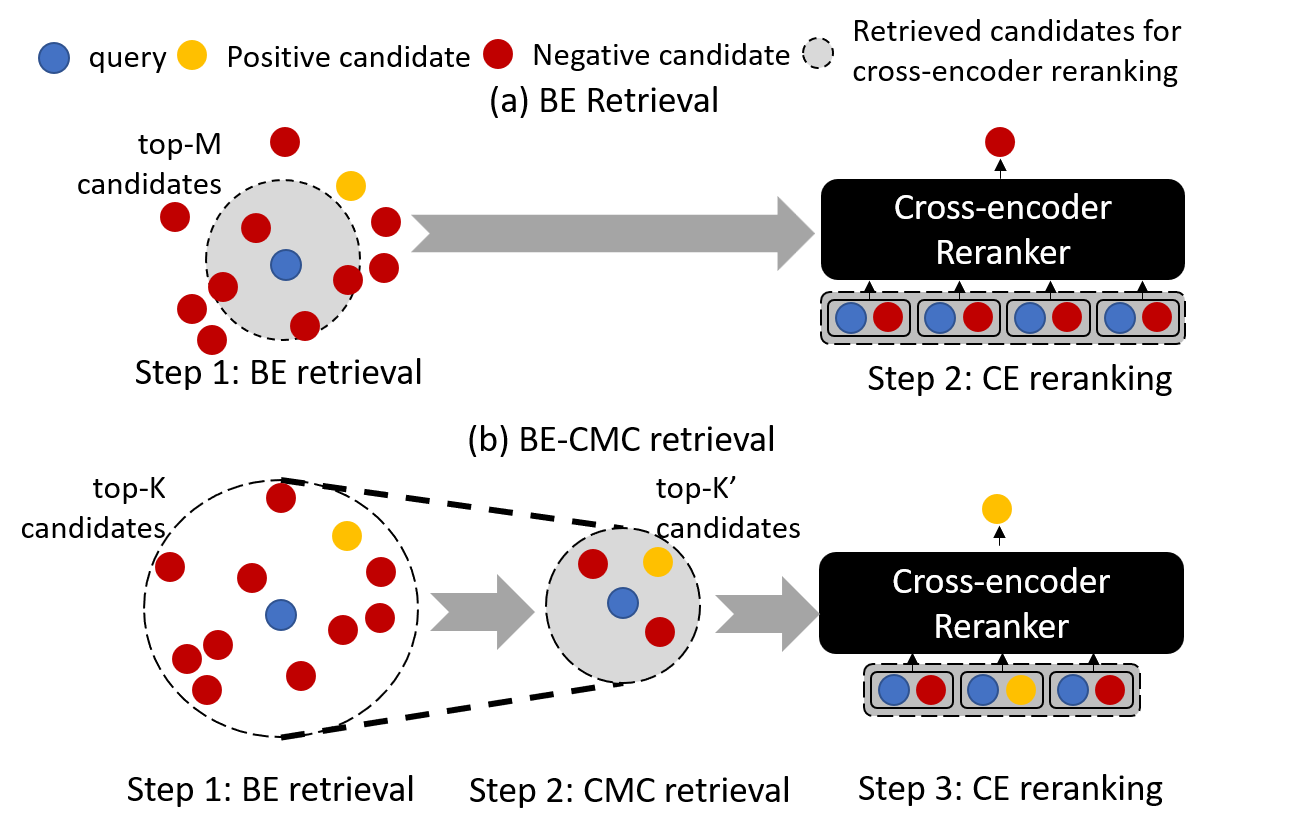}
  \caption{Illustration of candidate retrieval for cross-encoders (CE). Suppose cross-encoders can process up to M candidates due to limited scalability. (a) In bi-encoder (BE) retrieval, the BE-CE framework takes M candidates and risks missing the gold candidates due to inaccurate bi-encoders, causing the entire system to suffer from error propagation from the retriever and fail to get the correct candidate. (b) When \cmc\ is introduced as the seamless intermediate reranker (BE-CMC-CE), \cmc\ can consider a significantly larger pool (K) of BE candidates. This allows \cmc\ to provide much fewer K’ (K>M>K’) and higher-quality candidates to the CE while increasing the chance to include the positive candidate.}
  \label{fig:be-cmc-ce}
\end{figure}

We conduct two experiments on the ZeSHEL dataset to verify the impact of \cmc\ as the seamless intermediate retriever (BE+\cmc+CE). We examine whether the introduction of \cmc\ can improve retrieval performance with negligible overhead as promised. In the first experiment, we compare the performance and speed of \cmc\ plugged in with bi-encoders (BE+\cmc) with other retrieval pipelines. Remarkably, even when other rerankers are plugged in with the same bi-encoder, \cmc\ still achieves the highest Recall@k (Table \ref{tab:retriever}) at a marginal latency increase. In the second experiment, we assess how a more confident set of candidates retrieved by BE+\cmc\ contributes to improving end-to-end (BE+\cmc+CE) accuracy compared to solely using bi-encoders (Figure \ref{fig:be-cmc-ce}).

\paragraph{Baselines}  
To assess \cmc's effectiveness in enhancing retrieval, we evaluate BE+\cmc\ on 512 bi-encoder retrieved candidates and compare it to baselines categorized into two types: single- and multi-view retrievers.\footnote{Single-view retrievers consider only a single global view derived from the entire sentence, whereas multi-view retrievers divide candidate information into multiple local views.}  We use bi-encoders \cite{yadav2022efficient} and MVD \cite{liu2023towards} as the first-stage retrievers for the single-view and multi-view settings, respectively. For the baselines, we select the state-of-the-art retrievers for the ZeSHEL dataset. For single-view retrievers, we select the poly-encoder \cite{humeau2019poly}, Sum-of-max \cite{zhang2021understanding}, Arbo-EL \cite{agarwal2022zero}, and GER \cite{wu2023modeling}. Among these, Arbo-EL and GER utilize graph information, unlike \cmc\ and other baselines. For multi-view retrievers, we include MuVER \cite{ma2021muver} and MVD \cite{liu2023towards}.

\paragraph{Experimental Results}
In Table \ref{tab:retriever}, plugging in \cmc\ with a single-view retriever outperforms baselines across all $k$, demonstrating its effectiveness in the end-to-end retrieval process. With a marginal increase in latency (+0.07x), \cmc\ boosts recall@64 to 91.51\% on the candidates from the first-stage retriever, which has a recall@64 of 87.95\%. 
Especially, the recall of Poly-encoder and Sum-of-max lags behind \cmc\ even when they are plugged in with the same bi-encoders (BE+Poly \& BE+SOM). Sum-of-max, which closely follows \cmc, requires a tremendous index (60x of \cmc) to achieve comparable latency to \cmc. 
To show that \cmc\ seamlessly enhances any retriever type, we examine the increase in recall of \cmc\ upon a multi-view retriever (MVD+\cmc). The results show that \cmc\ consistently improves recall performance, moving from 91.55\% to 92.36\% at recall@64. This demonstrates \cmc's general capability to enhance recall performance, regardless of the first-stage retriever. For the effect of the number of candidates from the first-stage retriever, see \sref{sec:detailled-number}.

We question whether BE+\cmc\ can reduce the latency of the overall retrieval and reranking process while maintaining the overall accuracy (Figure \ref{fig:be-cmc-ce}). In essence, if we can have fewer but higher quality candidates, end-to-end accuracy can be improved while CE forward passes are called fewer times with a reduced set of candidates. To examine the quality of candidates from the seamless intermediate reranker \cmc, we report the final reranking accuracy of cross-encoders when candidates are retrieved by \biint+\cmc\ and compare it to conventional BE retrieval (Table \ref{tab:retriever-ce}).

Table \ref{tab:retriever-ce} shows that cross-encoders outperform conventional bi-encoders, even with fewer candidates retrieved by \cmc. Cross-encoders with 16 candidates from \cmc\ are 1.75x faster and achieve slightly better accuracy compared to using 64 bi-encoder candidates (line 3 vs. 8-9). Furthermore, cross-encoders reach the best accuracy with 64 candidates from \cmc, surpassing the accuracy obtained with the same number of bi-encoder candidates, with only a marginal increase in latency (line 3 vs. 11).




\begin{table}[t!]
\centering
\resizebox{\columnwidth}{!}{%
\begin{tabular}{lll|l|lllll|l}
\hline
 &\multicolumn{2}{l|}{Retrieved (k)} & Recall@k & \multicolumn{5}{l|}{Unnormalized Accuracy}                                                                                                                                                                  & Comparative                                                     \\
&Bi-encoder          & \cmc         &          & \begin{tabular}[c]{@{}l@{}}Forgotten \\ Realms\end{tabular} & Lego                       & \begin{tabular}[c]{@{}l@{}}Start \\ Trek\end{tabular} & \multicolumn{1}{l|}{Yugioh} & \begin{tabular}[c]{@{}l@{}}Macro \\ Avg.\end{tabular}                    & \begin{tabular}[c]{@{}l@{}}Latency \\   (\%)\end{tabular} \\ \hline
 1&8                   & -            & 77.72    & 78.92                                                       & 65.14                      & 62.76                                                 & \multicolumn{1}{l|}{48.64}  & 63.87                      & 38.90\%                                                   \\
 2&16                  & -            & 81.52    & 80.17                                                       & 66.14                      & 63.69                                                 & \multicolumn{1}{l|}{49.64}  & 64.91                      & 48.85\%                                                   \\
 3&64                  & -            & 87.95    & 80.83                                                       & 67.81                      & 64.23                                                 & \multicolumn{1}{l|}{50.62}  & 65.87                      &              \cellcolor[HTML]{C0C0C0}100\%                                            \\ \hline
 4&64                  & 8            & 82.45    & 80.67                                                       & 66.56                      & \underline{64.54}                                     & \multicolumn{1}{l|}{50.71}  & 65.62                      & 43.04\%                                        \\
 5&256                 & 8            & 82.86    & \underline{80.92}                                           & 66.89                      & \underline{64.42}                                     & \multicolumn{1}{l|}{50.86}  & 65.77                      &               43.36\%                                     \\
 6&512                 & 8            & 82.91    & 80.75                                                       & 67.14                      & \underline{64.35}                                     & \multicolumn{1}{l|}{51.01}  & 65.81                      &                   43.55\%                                 \\ \hline
 7&64                  & 16           & 85.46    & 80.5                                                        & 66.97                      & \underline{64.47}                                     & \multicolumn{1}{l|}{50.68}  & 65.66                      &                         56.76\%                           \\
 8&256                 & 16           & 86.22    & 80.75                                                       & 67.31                      & \underline{\textbf{64.63}}                            & \multicolumn{1}{l|}{51.1}   & \underline{65.95}          & 57.08\%                                              \\
 9&512                 & 16           & 86.22    & \underline{80.83}                                           & 67.64                      & \underline{64.49}                                     & \multicolumn{1}{l|}{50.95}  & \underline{65.98}          & 57.27\%           \\ \hline
 10&256                 & 64           & \underline{90.91}    & \underline{\textbf{81.17}}                                  & 67.64                      & \underline{64.37}                                     & \multicolumn{1}{l|}{50.92}  & \underline{66.03}          & 104.46\%                                                     \\
 11&512                 & 64           & \textbf{91.51    }& \underline{81.00}                                              & \underline{\textbf{67.89}} & \underline{64.42}                                     & \multicolumn{1}{l|}{50.86}  & \underline{\textbf{66.04}} &    104.65\%                                         \\ \hline
\end{tabular}%
}
\caption{Unnormalized accuracy\tablefootnote{The unnormalized accuracy of the reranker in ZeSHEL is defined as the accuracy computed on the entire test set. In contrast, the normalized accuracy is evaluated on the subset of test instances for which the gold entity is among the top-k candidates retrieved by the initial retriever. For example, if the retriever correctly identifies candidates for three out of five instances and the reranker identifies one correct candidate, unnormalized accuracy is 1/5 = 20\%, and normalized accuracy is 1/3 = 33\%.} of cross-encoders across various candidate configurations on the ZeSHEL dataset. We \underline{underlined} when the cross-encoders show superior accuracy with candidates filtered by \cmc\ compared to those from bi-encoders. The top-performing scenarios in each category are highlighted in \textbf{bold}. We measure the comparative latency required for running cross-encoders over 64 bi-encoder candidates (260.84ms). For your reference, the \cmc\ runtime 2x when increasing the number of candidates by 16x (from 128 to 1048), while able to compare up to 16k candidates at once. (\sref{sec:app-runtime})}
\label{tab:retriever-ce}
\end{table}

\begin{table*}[!htbp]
\centering
\resizebox{0.9\textwidth}{!}{%
\begin{tabular}{ll|llllllll}
\hline
 &
  Tasks &
  \multicolumn{2}{l}{Entity Linking} &
  \multicolumn{2}{l}{Passage Ranking} &
  \multicolumn{2}{c}{Dialogue Ranking}&
  \multicolumn{2}{l}{Compuational Efficiency} \\
 &
  Datasets &
  \multicolumn{1}{l}{Wikipedia} &
  \multicolumn{1}{l}{ZeSHEL} &
  \multicolumn{2}{l}{MS MARCO Dev} &
  \multicolumn{2}{l}{DSTC7 Challenge}&
  Total Speed&
  Extra Memory\\
 &
   &
  \multicolumn{1}{l}{Accuracy} &
  \multicolumn{1}{l}{Accuracy} &
  R@1&
  MRR@10&
  \multicolumn{1}{l}{R@1} &
  \multicolumn{1}{l}{MRR@10}   &
   &
   \\ \hline
High-latency  & Cross-encoder & 80.2\small $\pm 0.2$& \textbf{65.9$^\dag$}&  \textbf{25.4}&  \textbf{36.8}& 64.7       & 73.2& 12.9x& -    \\ \hline
Medium-latency& Deformer      & 79.6\small $\pm$0.8& \underline{63.6}\small$\pm 0.3$&  23.0$^\dag$&  35.7$^\dag$& \textbf{68.6}& \textbf{76.4}& 4.39x& 125x \\
              & Sum-of-max    & \multirow{2}{*}{\underline{80.7}\small$\pm 0.2$}                   & \multirow{2}{*}{58.8\small$\pm 1.0$}&  \multirow{2}{*}{22.8}$^\dag$&  \multirow{2}{*}{35.4}$^\dag$& \multirow{2}{*}{66.9}         & \multirow{2}{*}{75.5}& 5.20x & -  \\
              & - w/ offline indexing&  & &  &  &         & & 1.05x & 125x \\\hline
Low-latency& Bi-encoder    & 77.1$^\dag$                        & 52.9$^\dag$          & 22.9 &  35.3& 67.8& 75.1& 1x    & 1x   \\
              & Poly-encoder  & 80.2\small$\pm0.1$& 57.6\small$\pm 0.6$& 23.5 &  35.8& \underline{68.6}& \underline{76.3}& 1.01x & 1.0x \\
              & MixEncoder    & \multicolumn{1}{l}{75.4\small$\pm1.4$} & 57.9\small$\pm 0.3$&  20.7$^\dag$&  32.5$^\dag$& 68.2$^\dag$& 75.8$^\dag$& 1.12x & 1.0x \\ 
 &
  CMC (Ours) &
  \textbf{80.9}\small$\pm 0.1$&
  59.2\small$\pm 0.3$ &
  \multicolumn{1}{l}{\underline{23.9}} &
  \multicolumn{1}{l}{\underline{35.9}} &
  68.0&
  75.7&
  1.17x &
  1.0x \\ \hline
\end{tabular}%
}
\caption{Reranking Performance on four datasets with three downstream tasks: Entity Linking (Wikipedia-KB based datasets \cite{hoffart2011robust, guo2018robust, cucerzan2007large}, ZeSHEL \cite{logeswaran2019zero}, Passage Ranking (MS MARCO Passage Ranking \cite{nguyen2016ms}, and Dialogue Ranking \cite{gunasekara2019dstc7}. The best result is denoted in \textbf{bold} and the second-best result is \underline{underlined}. MRR stands for mean reciprocal rank. In the entity linking datasets, the results are averaged across five random seeds. To show the computing resources required for the reranking process, we define reranking latency in terms of relative latency and additional memory usage compared to bi-encoders.   $^\dag$ indicates that more favorable results are sourced from \citet{wu2020scalable, yang-etal-2023-enough, yadav2022efficient}, respectively. }
\label{tab:ranking}
\end{table*}

\subsection{CMC as the Final Stage Reranker}
\label{subsec:reranker}
\paragraph{Baselines}
Baselines are categorized into high-, medium-, and low-latency models. 
We adopt cross-encoders as our primary baseline for the high-latency model. 
For the medium-latency models, we include Deformer and Sum-of-max, which utilize all token embeddings to represent candidate information. For the low-latency models, we include the Bi-encoder, Poly-encoder, and Mixencoder, all of which require a single vector embedding for representation and have a serving time similar to that of the Bi-encoder. In this context, \cmc\ is classified as a low-latency method because it requires a single embedding for the candidate and takes 1.17x serving time of the Bi-encoder.  

 \paragraph{Comparison with Low-latency Models}
\cmc\ is highly effective across diverse datasets, outperforming or being comparable to other low-latency baselines. Notably, \cmc\ surpasses bi-encoders on every dataset with only a marginal increase in latency. This indicates that replacing simple dot products with self-attention layers across multiple candidates can enhance reranking performance, likely by taking advantage of the relational dynamics among the candidates. Evaluated against the Poly-encoder and MixEncoder, \cmc\ demonstrates superior prediction capability in tasks like passage ranking and entity linking, which require advanced reading comprehension capability. 

 \paragraph{Comparison with Medium-latency Models}
When compared with Medium-latency models such as Deformer and Sum-of-max, \cmc\ demonstrates its capability not only in memory efficiency but also in maintaining strong performance. \cmc\ mostly surpasses these models in entity linking and passage ranking tasks. Also, \cmc\ offers significant improvements in speed over Deformer (1.17x vs. 4.39x) and Sum-of-max without caching (1.17x vs. 5.20x). 
For Sum-of-max with caching, it requires a huge memory index size (125x) to accomplish a similar latency to \cmc. If a 125x index size is not available in practice, the speed becomes impractical introducing scalability limitations.
This analysis implies that \cmc's single-vector approach is significantly faster and more memory efficient, while still demonstrating a comparable ability to represent candidate information with fewer tokens, often surpassing more complex methods.

\paragraph{Comparison with High-latency Models}
Given the importance of computational resources and serving time in applications, \cmc\ is a practical alternative to cross-encoders, with 11.02x speedup and comparable reranking accuracy. \cmc\ outperforms the cross-encoder in the Wikipedia entity linking (+0.7p accuracy) and DSTC7 dialogue ranking (+3.3p MRR). Also, \cmc\ presents a competitive result in MS MARCO and ZeSHEL dataset, achieving the second- or third-best in prediction. This comparison suggests that the self-attention layer in \cmc\ effectively substitutes for the token-by-token interaction in cross-encoders while enhancing the computational efficiency of the reranking process. 
\begin{figure}
\centering
    \includegraphics[width=\columnwidth]{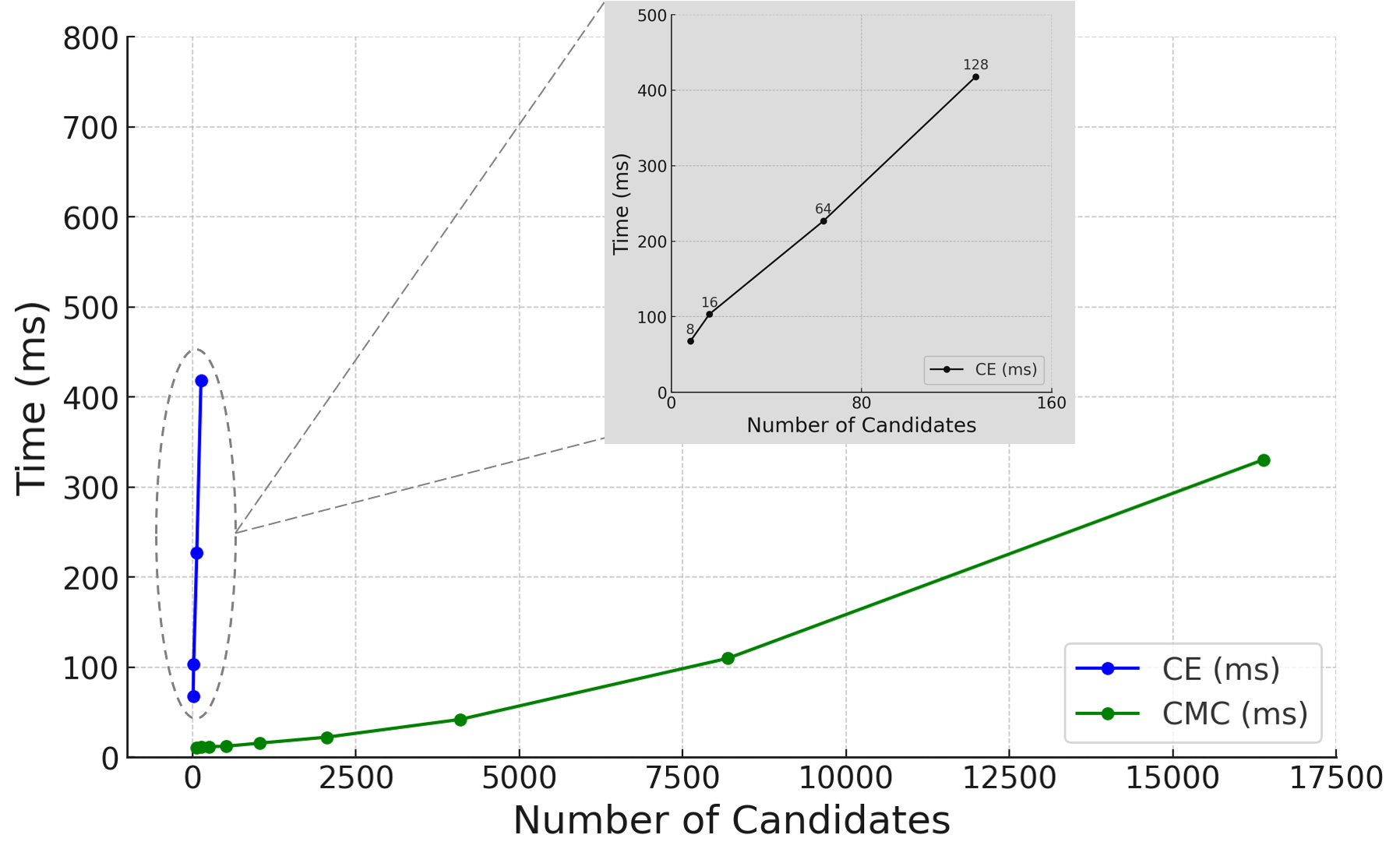}
\caption{The relationship between the number of candidates and the corresponding time measurements in milliseconds for two different models: Cross-encoder (CE) and Comparing Multiple Candidates (\cmc).}
    \label{fig:runtime}
\end{figure}

In summary, to achieve the best accuracy, we recommend the 3-stage retrieval pipeline of bi-encoders + \cmc\ + cross-encoders (BE-CMC-CE) that is both more accurate and substantially faster than the widely adopted bi-encoder + cross-encoder (BE-CE), as shown in Table \ref{tab:retriever-ce} and \sref{subsec:retriever}. If there’s a time constraint, using \cmc\ as the final reranker can be a good option since inferring with 16 candidates using a cross-encoder takes approximately the same amount of time as comparing around 10K candidates with \cmc\ (Figure \ref{fig:runtime})

\subsection{Ablation Study}
\label{subsec:ablation}
Through the experiments, we notice an improved reranking performance on \cmc\ when transferring the sentence encoder from another domain. To examine whether this is \cmc-specific characteristic, we conduct an experiment that investigate how different sentence encoder initializations affect the performance of late-interaction models. 
For each model, we consider sentence encoder initializations with BERT-based bi-encoders fine-tuned for an in-domain (ZeSHEL; \cite{yadav2022efficient}) and out-domain (MS-MARCO; \cite{guo2018robust}), as well as vanilla BERT \cite{devlin2018bert}; then for each combination of model and sentence-encoder initialization, we fine-tune the model on ZeSHEL dataset and report its test set results.

In Table \ref{tab:ablation2}, different initialization strategies show different effects for each model. \cmc\ and Poly-encoder show significant performance increases with out-of-domain sentence encoder initialization. This can be attributed to both models utilizing single candidate embeddings. Other models, such as Sum-of-max and MixEncoder, show negligible impact from sentence encoder initialization, whereas Deformer and Bi-encoder perform best with vanilla BERT. These findings suggest that  \cmc\ and the poly-encoder, which compress candidate information into single embeddings, can benefit from initialization from out-of-domain sentence encoders. As a practical recommendation, we advise practitioners to try out-of-domain initialization when using \cmc\ for potentially improved performance.

\begin{table}[!htbp]
\resizebox{\columnwidth}{!}{%
\begin{tabular}{lcllc}
\hline
              (Valid/Test) &         & \multicolumn{3}{c}{Sentence Encoder Initialization}                                                                                                                                                                             \\ \cline{3-5} 
              & \multicolumn{1}{l}{} & \multirow{2}{*}{\begin{tabular}[c]{@{}l@{}}Vanilla\\ BERT\end{tabular}} & \multicolumn{2}{c}{Fine-tuned with}                                                                                                                   \\ \cline{4-5} 
              & Model                &                                                                         & \multicolumn{1}{c}{\begin{tabular}[c]{@{}c@{}}In-domain\\ (ZeSHEL)\end{tabular}} & \begin{tabular}[c]{@{}c@{}}Out-of-domain\\ (MS MARCO)\end{tabular} \\ \hline
Medium- & Deformer             & \textbf{65.40/63.58}                                                    & 64.42/62.43                                                                      & 57.01/57.46                                                        \\
Latency       & Sum-of-max           & \multicolumn{1}{c}{\textbf{59.57}/58.37}                                         & 58.77/57.65                                                                      & 59.15/\textbf{58.79}                                                        \\ \hline
Low-          & Bi-encoder           & \textbf{55.54/52.94}                                                             & 55.54/52.94                                                                      & 49.32/44.01                                                        \\
Latency       & Poly-encoder         & 53.37/52.49                                                             & 55.75/54.22                                                                      & \textbf{57.41/58.22}                                                        \\
              & MixEncoder           & \textbf{58.63/57.92      }                                                       & 58.32/57.68                                                                      & 58.52/57.70                                                        \\
              & CMC (Ours)           & 56.15/55.34                                                             & 58.04/56.20                                                                      & \textbf{60.05/59.23 }                                                       \\ \hline
\end{tabular}
}
\caption{Comparison of unnormalized accuracy on valid/test set of ZeSHEL over different sentence encoder initialization (Vanilla BERT \cite{devlin2018bert}, Bi-encoder fine-tuned for in- \cite{yadav2022efficient} and out-of-domain \cite{guo2020accelerating}) dataset. We denote the best case for each method as bold.}
\label{tab:ablation2}
\end{table}

\section{Conclusion}
In this paper, we present a novel and intuitive retrieval and reranking framework, Comparing Multiple Candidates (\cmc). By contextualizing the representations of candidates through the self-attention layer, \cmc\ achieves improvements in prediction performance with a marginal increase in speed and memory efficiency. Experimental results show that \cmc\ acts as a seamless intermediate reranker between bi-encoders and cross-encoders. The retrieval pipeline of BE-CMC-CE is not only more accurate but also substantially faster than the widely adopted bi-encoder + cross-encoder (BE-CE). Meanwhile, experiments on four different datasets demonstrate that \cmc\ can serve as the efficient final stage reranker. These empirical results emphasize \cmc's effectiveness, marking it as a promising advancement in the field of neural retrieval and reranking.
\section*{Limitations}
In the future, we plan to test the \cmc's performance with over 1000 candidates with batch processing. It has not yet been extensively researched whether \cmc\ can effectively retrieve from a large collection, e.g., a collection comprising more than 1 million candidates. Furthermore, we plan to tackle the issue that arises from the concurrent operation of both a bi-encoder and \cmc\ index, which currently requires double the index size. This is a consequence of running two separate encoder models in parallel. To address this, we will investigate an end-to-end training scheme, thereby enhancing the practicality and efficiency of running both the Bi-encoder and \cmc\ simultaneously.
\section*{Acknowledgement}
We thank Nishant Yadav for his helpful discussions and feedback. This work was supported in part by the National Research Foundation of Korea (NRF) grant (RS-2023-00280883, RS-2023-00222663) and New Faculty Startup Fund from Seoul National University, and with the aid of computing resources from Artificial Intelligence Industry Center Agency, Google cloud platform research credits, and the National Super computing Center with super computing resources including technical support (KSC-2023-CRE-0176).


\newpage
\appendix
\section{Potential Risks}
This research examines methods to accelerate the retrieval and reranking process using efficient and effective \cmc. While the proposed \cmc\ might exhibit specific biases and error patterns, we do not address these biases in this study. It remains uncertain how these biases might affect our predictions, an issue we plan to explore in future research.
\section{Detailed Information of Datasets}
\label{sec:stat-datasets}
\paragraph{Wikipedia Entity Linking} For standard entity linking, we use AIDA-CoNLL dataset \cite{hoffart2011robust} for in-domain evaluation, and WNED-CWEB \cite{guo2018robust} and MSNBC \cite{cucerzan2007large} datasets for out-of-domain evaluation. These datasets share the same Wikipedia knowledge base. For comparison with the baseline results from \citet{wu2020scalable}, we employ the 2019 English Wikipedia dump, containing 5.9M entities. We employed a bi-encoder as an initial retriever that yields an average unnormalized accuracy of 77.09 and a recall@10 of 89.21. 
Unnormalized accuracy is measured for each dataset and macro-averaged for test sets.

AIDA-CoNLL dataset is licensed under a Creative Commons Attribution-ShareAlike 3.0 Unported License. We are not able to find any license information about WNED-CWEB and MSNBC datasets.

\paragraph{Zero-shot Entity Linking (ZeSHEL)} ZeSHEL \cite{logeswaran2019zero} contains mutually exclusive entity sets between training and test data.
The dataset comprises context sentences (queries) each containing a mention linked to a corresponding gold entity description within Wikia knowledge base. Unlike Wikipedia entity linking datasets where the entity set of train and test set overlaps, the entity set for ZeSHEL is mutually exclusive and this setup tests the model's ability to generalize to new entities. We employed a bi-encoder from \cite{yadav2022efficient} whose recall@64 is 87.95. On top of these candidate sets, we report macro-averaged unnormalized accuracy, which is calculated for those mention sets that are successfully retrieved by the retriever and macro-averaged across a set of entity domains. For statistics of entity linking datasets, see Table \ref{table:stat-zeshel}. ZeSHEL is licensed under the Creative Commons Attribution-Share Alike License (CC-BY-SA).

The predominant approach for reranking in ZeSHEL dataset is based on top-64 candidate sets from official BM25 \cite{logeswaran2019zero} or bi-encoder \cite{wu2020scalable, yadav2022efficient}. On top of these candidate sets, we report macro-averaged normalized accuracy, which is calculated for those mention sets that are successfully retrieved by the retriever and macro-averaged across a set of entity domains.

\begin{table}[h!]
\resizebox{\columnwidth}{!}{%
\begin{tabular}{lllc}
\hline
\multicolumn{2}{c}{Dataset}         & \# of Mentions & \multicolumn{1}{l}{\# of Entities} \\ \hline
AIDA                    & Train     & 18848          & \multirow{5}{*}{5903530}           \\
                        & Valid (A) & 4791           &                                    \\
                        & Valid (B) & 4485           &                                    \\ \cline{1-2}
\multicolumn{2}{l}{MSNBC}           & 656            &                                    \\
\multicolumn{2}{l}{WNED-WIKI}       & 6821           &                                    \\ \hline
\multirow{3}{*}{ZeSHEL} & Train     & 49275          & \multicolumn{1}{l}{332632}         \\
                        & Valid     & 10000          & \multicolumn{1}{l}{89549}          \\
                        & Test      & 10000          & \multicolumn{1}{l}{70140}          \\ \hline
\end{tabular}
}
\caption{Staistics of Entity Linking datasets.}
\label{table:stat-zeshel}
\end{table}

\paragraph{MS MARCO} We use a popular passage ranking dataset MS MARCO which consists of 8.8 million web page passages. MS MARCO originates from Bing's question-answering dataset with pairs of queries and passages, the latter marked as relevant if it includes the answer. Each query is associated with one or more relevant documents, but the dataset does not explicitly denote irrelevant ones, leading to the potential risk of false negatives. For evaluation, models are fine-tuned with approximately 500K training queries, and MRR@10, Recall@1 are used as a metric. To compare our model with other baselines, we employed Anserini's BM25 as a retriever \cite{nogueira2019document}. The dataset is licensed under Creative Commons Attribution 4.0 International.

\paragraph{DSTC 7 Challenge (Track 1)}
For conversation ranking datasets, we involve The DSTC7 challenge (Track 1) \cite{yoshino2019dialog} . DSTC 7 involves dialogues taken from Ubuntu chat records, in which one participant seeks technical assistance for diverse Ubuntu-related issues. For these datasets, an official candidate set which includes gold is provided. For statistics for MS MARCO and DSTC 7 Challenge, see Table \ref{table:stat-marco}
\begin{table}[h!]
\resizebox{\columnwidth}{!}{%
\begin{tabular}{lllll}
\hline
Datasets  & Train   & Valid & Test  & \# of Candidates \\
          &         &       &       & per Query        \\ \hline
MS MARCO  & 498970  & 6898  & 6837  & 1000             \\
DSTC 7    & 100000  & 10000 & 5000  & 100              \\ \hline
\end{tabular}}
\caption{Statistics of MS MARCO \& Conversation Ranking Datasets.}
\label{table:stat-marco}
\end{table}



\begin{table*}[!htbp]
\resizebox{\textwidth}{!}{%
\begin{tabular}{l|ll|l|l}
\hline
                     & \multicolumn{2}{l|}{Entity Linking}                                                         & Passage Ranking                              & Dialogue Ranking\\
                     & AIDA-train                                   & ZeSHEL                                       & MS MARCO                                     & DSTC7                                        \\ \hline
max. query length    & 32                                           & 128                                          & 32                                           & 512                                          \\
max. document length & 128                                          & 128                                          & 128                                          & 512                                          \\
learning rate        & \{\textbf{1e-5},5e-6,2e-6\} & \{\textbf{1e-5},2e-5,5e-5\} & \{1e-5,5e-6,\textbf{2e-6}\} & \{\textbf{1e-5},2e-5,5e-5\} \\
batch size           & 4                                            & 4                                            & 8                                            & 8                                            \\
hard negatives ratio & 0.5                                          & 0.5                                          & 1                                            & -                                            \\
\# of negatives      & 63                                           & 63                                           & 63                                           &                                              7\\
training epochs      & 4                                            & 5                                            & 3                                            & 10                                           \\ \hline
\end{tabular}
}
\caption{Hyperparameters for each dataset. We perform a grid search on learning rate and the best-performing learning rate is indicated as bold.}
\label{tab:hyperparams}
\end{table*}
\section{Training Details}
\label{sec:detail-training}
\paragraph{Negative Sampling} Most of previous studies that train reranker \cite{wu2020scalable, xu2023read} employ a fixed set of top-\(k\) candidates from the retriever. In contrast, our approach adopts hard negative sampling, a technique derived from studies focused on training retrievers \cite{zhang2021understanding}. Some negative candidates are sampled based on the retriever's scoring for query-candidate pair $(q, c_{q,j})$:
\begin{equation}
\begin{aligned}
&\forall j \in \{1, \ldots, K\} \setminus \{\text{gold index}\}, \\
&\quad \tilde{c}_{q, j} \sim \frac{\exp(s_{\text{retriever}}(q, \tilde{c}_{q, j}))}{\sum_{\substack{k=1 \\ k \neq \text{gold index}}}^K \exp(s_{\text{retriever}}(q, \tilde{c}_{q, k}))}
\end{aligned}
\label{eq:sampling}
\end{equation}

To provide competitive and diverse negatives for the reranker, p\% of the negatives are fixed as the top-\( k \) negatives, while the others are sampled following the score distribution. 

As detailed in Table \ref{tab:hyperparams}, we implement a hard negative mining strategy for training \cmc\ and comparable baseline methods. Specifically, for the MS MARCO dataset, hard negatives are defined as the top 63 negatives derived from the CoCondenser model, as outlined in \citet{gao2022unsupervised}. In the case of entity linking datasets, we adhere to the approach established by \citet{zhang2021understanding}, where hard negatives are selected from the top 1024 candidates generated by a bi-encoder. Meanwhile, for dialogue ranking datasets, we do not employ hard negative mining, owing to the absence of candidate pool within these datasets.


\paragraph{Sentence Encoder Initialization} 
\label{subsec:init}
The initial starting point for both the query and candidate encoders can significantly impact performance. The sentence encoders for late interaction models including \cmc\ are initialized using either vanilla huggingface BERT \cite{devlin2018bert} or other BERT-based, fine-tuned models. These models include those fine-tuned on the Wikipedia dataset (BLINK-bi-encoder; \citet{wu2020scalable}) or MS MARCO (Cocondenser; \citet{gao2022unsupervised}). As the cross-encoder is the only model without sentence encoder, we initialize cross-encoder using pre-trained BERT (BLINK-cross-encoder; \citet{wu2020scalable}) or vanilla BERT. 

We initialize the sentence encoder for \cmc\ and other baselines using (1) vanilla BERT and (2) the BLINK bi-encoder for Wikipedia entity linking datasets, and the MS-MARCO fine-tuned Cocondenser for other datasets. After conducting experiments with both starting points, we selected the best result among them. If more favorable results for baselines are found from prior works that conduct reranking over the same candidates, we sourced the numbers from these works.

\paragraph{Optimization} 
Our model employs multi-class cross-entropy as the loss function, regularized by Kullback-Leibler (KL) divergence between the reranker's scores and the retriever's scores. The loss function is formulated as follows:
\begin{equation}
\begin{split}
\mathcal{L}(q, \tilde{C}_q) &= -\lambda_1\sum_{i=1}^{K} y_i \log(p_i) \\
&+ \lambda_2 \sum_{i=1}^{K} p_i \log\left(\frac{p_i}{r_i}\right)
\end{split}
\end{equation}


For the query \( q\), \( y_i \) represents the ground truth label for each candidate \( \tilde{c}_{q,i} \),  \( p_i \) is the predicted probability for candidate \( \tilde{c}_{q,i} \) derived from the score function \( s_\theta \), \( r_i \) is the probability of the same candidate from the retriever's distribution, and \(\lambda_1\) and \(\lambda_2\) are coefficients forming a convex combination of the two losses.

\paragraph{Extra Skip Connection} \cmc\ is trained end-to-end, where the self-attention layer is trained concurrently with the query and candidate encoders. In addition to the inherent skip connections present in the transformer encoder, we have introduced an extra skip connection following \citet{he2016deep} to address the vanishing gradient problem commonly encountered in deeper network layers. Specifically, for an encoder layer consisting of self-attention layer $\mathcal{F}(\mathbf{x})$, the output is now formulated as $\mathbf{x} + \mathcal{F}(\mathbf{x})$, with $\mathbf{x}$ being the input embedding. This training strategy ensures a more effective gradient flow during backpropagation, thereby improving the training stability and performance of our model.

\begin{table*}[]
\centering
\resizebox{\textwidth}{!}{
\begin{tabular}{l|llllll|ll}
\hline
                                                                   & \multicolumn{6}{l|}{Test}                     & \multicolumn{2}{l}{Valid} \\
Method                                                             & R@1   & R@4   & R@8   & R@16  & R@32  & R@64  & R@1         & R@64        \\ \hline
Bi-encoder & 52.94 & 64.51 & 71.94 & 81.52 & 84.98 & 87.95 & 55.45       & 92.04       \\
BE + \cmc (64)                                  & \textbf{59.22} & \textbf{77.69} & 82.45 & 85.46 & 87.28 & 87.95 & \textbf{60.27 }      & 92.04       \\
BE + \cmc (128)                                 & 59.13 & \underline{77.65} & 82.72 & 85.84 & 88.29 & 89.83 & \underline{60.24}       & 93.22       \\
BE + \cmc (256)                                 & \underline{59.13} & 77.6  & \underline{82.86} & \underline{86.21} & \underline{88.96} & \underline{90.93} & 60.13       & \underline{93.63}       \\
BE + \cmc (512)                                 & 59.08 & 77.58 & \textbf{82.91} & \textbf{86.32} & \textbf{89.33} & \textbf{91.51} & 60.1        & \textbf{93.89}       \\ \hline
\end{tabular}}
\caption{Retrieval performance by the number of candidates from the initial retriever. The numbers in parentheses (e.g., 128 for \cmc(128)) indicate the number of candidates which \cmc\ compares, initially retrieved by the bi-encoder. The best result is denoted in bold and the second-best result is underlined.  }
\label{tab:retrieve-appendix}
\end{table*}
\section{Additional Results and Analysis}

\subsection{Reranking Latency of \ce\ and \cmc}
\label{sec:app-runtime}
In Figure \ref{fig:runtime}, we present the plot of runtime against the number of candidates. For \cmc, the model can handle up to 16,384 candidates per query, which is comparable to the speed of \ce\ for running 64 candidates. Running more than 128 and 16,384 candidates cause memory error on GPU for \ce\ and \cmc, respectively. \subsection{Effect of Number of Candidates on Retrieval Performance}
\label{sec:detailled-number}
In Table \ref{tab:retrieve-appendix}, we present detailed results of retrieval performance on varying numbers of candidates from the initial bi-encoder. Recall@k increased with a higher number of candidates. It indicates that \cmc\ enables the retrieval of gold instances that could not be retrieved by a bi-encoder, which prevents error propagation from the retriever. It is also noteworthy that \cmc, which was trained using 64 candidates, demonstrates the capacity to effectively process and infer from a larger candidate pool (256 and 512) while giving an increase in recall@64 from 82.45 to 82.91.

\subsection{Detailed Information of Entity Linking Performance}
\label{sec:detailled-retrieve}
In Table \ref{tab:wiki}, we present detailed results for each dataset in Wikipedia entity linking task. Also, in table \ref{tab:zeshel}, we present detailed results for each world in ZeSHEL test set.

\begin{table}[htbp]
\resizebox{\columnwidth}{!}{
\begin{tabular}{ll|llll|r}
\hline
 &Method                                               & Valid (A)& Test (B)& MSNBC*                             & \begin{tabular}[c]{@{}c@{}}WNED-\\CWEB*\end{tabular} & \multicolumn{1}{l}{Average}                                       \\ \hline
  High-&Cross-encoder & 82.12                              & 80.27                              & 85.09                              & \multicolumn{1}{l|}{68.25}&77.87                                                                              \\
   Latency &Cross-encoder $^\dag$ & 87.15                              & 83.96                              & 86.69                              & \multicolumn{1}{l|}{69.11}&80.22\\ \hline
 Intermediate-& Sum-of-max $^\dag$& \underline{90.84}& \textbf{85.30}& 86.07& \multicolumn{1}{l|}{\textbf{70.65}}&\underline{80.67}\\
  Latency& Deformer$^\dag$& 90.64& 84.57& 82.92& \multicolumn{1}{l|}{66.97}&78.16\\\hline
 Low-&Bi-encoder & 81.45                              & 79.51                              & 84.28                              & \multicolumn{1}{l|}{67.47}      & 77.09                                                           \\
  Latency&Poly-encoder$^\dag$                                  & 90.64& 84.79& \underline{86.30}& \multicolumn{1}{l|}{69.39}      & 80.16\\
 & MixEncoder$^\dag$& 89.92& 82.69& 78.24& \multicolumn{1}{l|}{64.00}&76.27\\ 
 &\cmc  $^\dag$                                                & \textbf{91.16}    & \underline{85.03}& \textbf{87.35}& \multicolumn{1}{l|}{\underline{70.34}}      & \textbf{80.91}\\ \hline
\end{tabular}}
\caption{Unnormalized accuracy on Wikipedia entity linking dataset (AIDA \cite{hoffart2011robust}, MSNBC \cite{cucerzan2007large}, and WNED-CWEB \cite{guo2018robust}). \textit{Average} means macro-averaged accuracy for three test sets. The best result is denoted in bold and the second best result is denoted as \underline{underlined}. * is out of domain dataset. $^\dag$ is our implementation.}
\label{tab:wiki}
\end{table}

\begin{table}[htbp]
\resizebox{\columnwidth}{!}{
\begin{tabular}{ll|llllll}
\hline
                                        && \multicolumn{1}{l|}{Valid}          & \multicolumn{5}{c}{Test (By Worlds)}                                                                                                                                                                                                                                                                                      \\
 &Method                                 & \multicolumn{1}{l|}{}    & \begin{tabular}[c]{@{}c@{}}Forgotten\\ Realms\end{tabular}& \multicolumn{1}{c}{Lego} & \multicolumn{1}{c}{Star Trek} & \multicolumn{1}{c|}{Yugioh} & \multicolumn{1}{c}{Avg.}              \\ \hline
 High-&Cross-encoder & \multicolumn{1}{l|}{67.41}          & 80.83& 67.81& 64.23& \multicolumn{1}{l|}{50.62}  & 65.87\\
 Latency&\begin{tabular}[c]{@{}l@{}} Cross-encoder \\ (w/ \cmc ) \end{tabular}                    & \multicolumn{1}{l|}{70.22} & 81.00& 67.89& 64.42& \multicolumn{1}{l|}{50.86}  & 66.04\\ \hline
 Intermediate-& Sum-of-max& \multicolumn{1}{l|}{59.15}& 73.45& 58.83& 57.63& \multicolumn{1}{l|}{45.29}&58.80\\
 Latency& Deformer& \multicolumn{1}{l|}{56.95}& 73.08& 56.98& 56.24& \multicolumn{1}{l|}{43.55}&57.46\\ \hline
 Low-& Bi-encoder& \multicolumn{1}{l|}{55.45}& 68.42& 51.29& 52.66& \multicolumn{1}{l|}{39.42}&52.95\\ 
 Latency&Poly-encoder& \multicolumn{1}{l|}{57.19}          & 71.95& 58.11& 56.19& \multicolumn{1}{l|}{43.60}  & 57.46\\
 &MixEncoder& \multicolumn{1}{l|}{58.64}          & 73.17& 56.29& 56.99& \multicolumn{1}{l|}{43.01}  & 57.36\\
 &\cmc (Ours)                             & \multicolumn{1}{l|}{60.05} & 73.92& 58.96& 58.08& \multicolumn{1}{l|}{45.69}  & 59.16\\ \hline
\end{tabular}%
}
\caption{Detailed Reranking Performance on Zero-shot Entity Linking (ZeSHEL) valid and test set \cite{logeswaran2019zero}. Macro-averaged unnormalized accuracy is measured for candidates from Bi-encoder \cite{yadav2022efficient}.The best result is denoted in \textbf{bold}.}
\label{tab:zeshel}
\end{table}
\subsection{Ranking Performance on ZeSHEL BM25 candidate sets}
In many previous works \cite{wu2020scalable, xu2023read}, the performance of models over BM25 candidates \cite{logeswaran2019zero} has been reported. In Table \ref{tab:zeshel-bm25}, we present the performance of \cmc\ to illustrate its positioning within this research landscape.

\begin{table}[htbp]
\resizebox{\columnwidth}{!}{
\begin{tabular}{l|llll|ll} \hline
Methods                                                 & \begin{tabular}[c]{@{}c@{}}Forgotten\\ Realms\end{tabular} & \multicolumn{1}{c}{Lego} & \multicolumn{1}{c}{Star Trek} & \multicolumn{1}{c}{Yugioh} & \multicolumn{1}{|c}{Macro Acc.} & \multicolumn{1}{c}{Micro Acc.} \\ \hline
Cross-encoder \cite{wu2020scalable} & 87.20                                & 75.26                    & 79.61                         & 69.56                      & 77.90                          & 77.07                          \\
ReS \cite{xu2023read}               & 88.10                                & 78.44                    & 81.69                         & 75.84                      & 81.02                          & 80.40                          \\ \hline
ExtEnD \cite{de2020autoregressive}                                                 & 79.62                                & 65.20                    & 73.21                         & 60.01                      & 69.51                          & 68.57                          \\
GENRE \cite{de2020autoregressive}                                                   & 55.20                                & 42.71                    & 55.76                         & 34.68                      & 47.09                          & 47.06                          \\
Poly-encoder$^{\dag}$                                            & 78.90                                & 64.47                    & 71.05                         & 56.25                      & 67.67                          & 66.81                          \\
Sum-of-max$^{\dag}$                                              & 83.20                                & 68.17                    & 73.14                         & 64.00                      & 72.12                          & 71.15                          \\
Comparing Multiple Candidates (Ours)                      & 83.20                                & 70.63                    & 75.75                         & 64.83                      & 73.35                          & 72.41                          \\ \hline
\end{tabular}
}
\caption{Test Normalized accuracy of  \cmc\ model over retrieved candidates from BM25.  $^*$ is reported from \citet{xu2023read}. $^{\dag}$
 is our implementation.}
\label{tab:zeshel-bm25}

\end{table}

\begin{table}[!hbt]
\centering
\resizebox{\columnwidth}{!}{
\begin{tabular}{l|lll}
\hline
                & \multicolumn{2}{l|}{w/ bi-encoder retriever}                             & \multicolumn{1}{l}{w/ BM25 retriever}                \\

                      Methods    & Valid           & \multicolumn{1}{l|}{Test}       & Test                \\ \hline
\cmc\                     & \underline{65.29}           & \multicolumn{1}{l|}{\textbf{66.83}}      & \textbf{73.10}                \\ \hline
w/o extra skip connection & 64.78           & \multicolumn{1}{l|}{66.44}      & 73.07               \\
w/o regularization        & 64.45           & \multicolumn{1}{l|}{66.31}      & 72.94               \\
w/o sampling            & \textbf{65.32 }          & \multicolumn{1}{l|}{\underline{66.46}}      & \underline{72.97}               \\ \hline
\end{tabular}%
}
\caption{Normalized Accuracy on ZeSHEL test set for various training strategies}
\label{tab:ablation-training}
\end{table}
\begin{table}[!t]
\centering
\small{
\resizebox{\columnwidth}{!}{%
\begin{tabular}{ll}
\hline
\# of candidates & \begin{tabular}[c]{@{}l@{}}Recall@1\\ (Unnormalized Accuracy)\end{tabular} \\ \hline
16               & 65.02                                                                      \\
64               & 65.87                                                                      \\
512              & 65.85                                                                      \\ \hline
\end{tabular}}}
\caption{Normalized Accuracy on ZeSHEL test set for various training strategies}
\label{tab:cross-num}
\end{table}

\subsection{Ablation Study on Training Strategies}
\label{subsec:ablation-training}
In Table \ref{tab:ablation-training}, we evaluated the impact of different training strategies on the \cmc 's reranking performance on the ZeSHEL test set. The removal of extra skip connections results in only a slight decrease ranging from 0.03 to 0.39 points in normalized accuracy. Also, to examine the effects of a bi-encoder retriever, we remove regularization from the loss. It leads to a performance drop but still shows higher performance than sum-of-max, the most powerful baseline in the low latency method. Lastly, we tried to find the influence of negative sampling by using fixed negatives instead of mixed negatives. The result shows a marginal decline in the test set, which might be due to the limited impact of random negatives in training \cmc.

\subsection{Reranking Performance of Cross-encoders for Various Number of Candidates}

\label{subsec:cross-num}
In Table \ref{tab:cross-num}, we evaluated the impact of the different number of candidates on the cross-encoder's reranking performance on the ZeSHEL test set with a candidate set from the bi-encoder retriever. Even with a larger number of candidates, the unnormalized accuracy of the cross-encoder does not increase. Although the number of candidates from the bi-encoder increases from 64 to 512, recall@1 decreases by 0.01 points. 
\textbf{}

\end{document}